\documentclass[letterpaper, 10 pt, conference]{ieeeconf}  

\IEEEoverridecommandlockouts                              

\usepackage{url}

\usepackage[pdftex]{graphicx}
\usepackage{pgf}
\usepackage{mathptmx}
\usepackage{amsmath}
\usepackage{amssymb} 
\usepackage{xcolor}
\usepackage{balance}
\usepackage{subfigure}
\usepackage[colorlinks=true,linkcolor=.,urlcolor=.]{hyperref}
\usepackage{siunitx} 
\usepackage[utf8]{inputenc}
\usepackage{wrapfig}
\usepackage{euro}

\graphicspath{{figures}}
\DeclareGraphicsExtensions{.pdf,.jpeg,.jpg,.png}
\usepackage{textcomp} 

\newcommand{\blue}[1]{{\color{blue} {#1}}} 

\newif\ifdraft
\drafttrue 

\ifdraft
\else
	\renewcommand{\blue}[1]{} 
\fi

\usepackage{fancyhdr}
\fancypagestyle{IEEECopyright}{
    \lfoot{\copyright2022 IEEE. Accepted at 2022 IEEE/RSJ International Conference on Intelligent Robots and Systems (IROS). \\ DOI: TBA} 
}

\hypersetup{pdftitle={A Virtual 2D Tactile Array for Soft Actuators Using Acoustic Sensing},
            pdfauthor={Vincent Wall and Oliver Brock},
            pdfkeywords={Soft Material Robotics,Tactile Sensing,Acoustic Sensing,Pneumatic Actuators,Braille Display},
            pdfsubject={Accepted at 2022 IEEE/RSJ International Conference on Intelligent Robots and Systems}
        }

\title{\LARGE \bf
	A Virtual 2D Tactile Array for Soft Actuators Using Acoustic Sensing
}

\author{Vincent Wall$^{1,2}$ \quad\quad Oliver Brock$^{1,2}$
	\thanks{\hspace{-1em}$^1$ Robotics and Biology Laboratory, Technische Universit\"{a}t Berlin, Germany
	\newline
	$^2$ Science of Intelligence, Research Cluster of Excellence, Berlin, Germany 
	\newline
	We gratefully acknowledge financial support by the Deutsche Forschungsgemeinschaft (DFG, German Research Foundation) under Germany's Excellence Strategy -- EXC 2002/1 ``Science of Intelligence'' -- project number 390523135 and the German Priority Program DFG-SPP 2100 ``Soft Material Robotic Systems''.
	We thank Prof.\ Dr.-Ing.\ Ennes Sarradj and Simon Jekosch of the Department of Engineering Acoustics, TU Berlin, for their insights.}
	}

\begin{document}
	
\maketitle
\thispagestyle{IEEECopyright}
\pagestyle{empty}

\begin{abstract}
\sisetup{detect-all = true} 
We create a virtual 2D tactile array for soft pneumatic actuators using embedded audio components. 
We detect contact-specific changes in sound modulation to infer tactile information.
We evaluate different sound representations and learning methods to detect even small contact variations. We demonstrate the acoustic tactile sensor array by the example of a PneuFlex actuator and use a Braille display to individually control the contact of 29~x~4 pins with the actuator's 90~x~10\,mm palmar surface. 
Evaluating the spatial resolution, the acoustic sensor localizes edges in x- and y-direction with a root-mean-square regression error of \SI{1.67}{mm} and \SI{0.0}{mm}, respectively. Even light contacts of a single Braille pin with a lifting force of \SI{0.17}{N} are measured with high accuracy.
Finally, we demonstrate the sensor's sensitivity to complex contact shapes by successfully reading the 26 letters of the Braille alphabet from a single display cell with a classification rate of \SI{88}{\percent}.
\end{abstract}

\section{Introduction}
\label{sec:introduction}
\sisetup{detect-all = true} 

We present a sensorization approach for measuring two-dimensional contact patterns on soft pneumatic actuators using embedded, off-the-shelf audio components. This approach builds upon our previously published ``Active Acoustic Sensor''~\cite{zoller_active_2020} and demonstrates how to apply the method to create a 2D tactile array on the actuator's surface, all without any changes to the sensor hardware itself.

In previous work, we demonstrated the surprising versatility of the acoustic sensing approach. With only an embedded microphone and speaker, we were able to measure a wide range of different actuator properties, including the contact location and force and even the temperature of the actuator~\cite{wall_passive_2022}. All these measurements are obtained using the same sensor hardware.
In this paper, we extend the acoustic sensing approach to a new application domain: two-dimensional tactile sensing. The two key features we focus on are the sensor's ability to measure contact as a two-dimensional tactile image, and the possibility to easily reconfigure in software the shape and size of the tactile pixels (taxels) without changing the hardware. 

Sensorization addresses the challenge of using soft robotic hands and actuators in the real world, where their inherent compliance makes modeling interaction behaviors infeasible. Tactile feedback allows handling the uncertainty of soft manipulation by observing contact and reacting accordingly. Especially in contact-rich tasks like in-hand manipulation~\cite{bhatt_surprisingly_2021}, a good estimate of the object position in the hand allows to select the appropriate control strategy. And since manipulation tasks inherently involve a lot of visual occlusions, it is much preferable to have embedded, onboard sensors.

\begin{figure}
\centering
\includegraphics[width=1\linewidth]{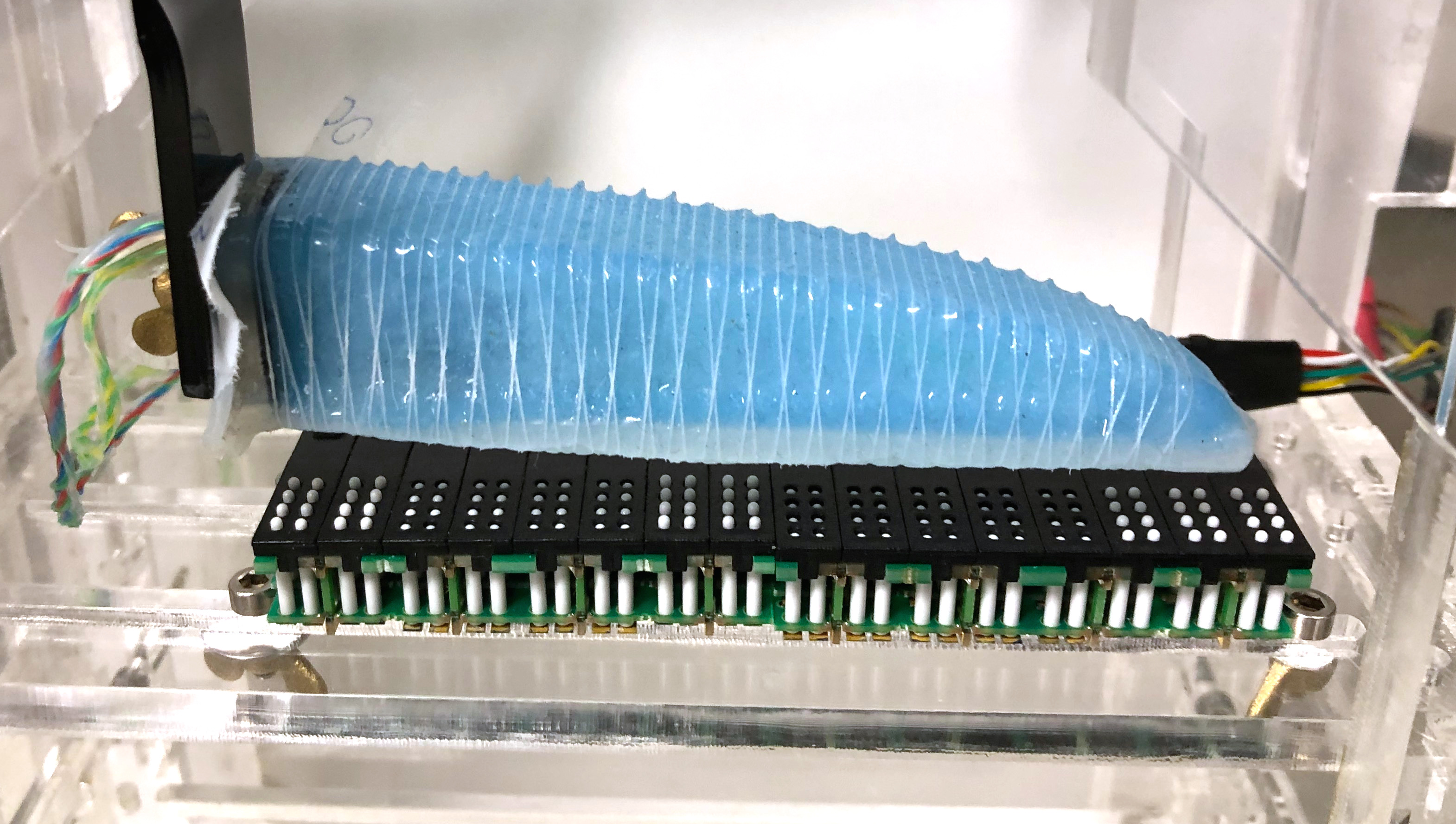}
\caption{We demonstrate our virtual, two-dimensional tactile array on a PneuFlex actuator with a Braille display. Using acoustic sensing, we measure even small changes in contact patterns created by the 32~x~4 pins.}
\label{fig:braillesetup}
\end{figure}

We demonstrate the acoustic tactile sensing approach on the PneuFlex actuator~\cite{deimel_novel_2016} using a modular Braille display to create various contact patterns (Fig.~\ref{fig:braillesetup}). We explain how to adapt the sensing method to handle the small differences in recorded sounds to measure fine contact details. As a result, we can show that the acoustic tactile array achieves a 2D contact location accuracy of \SI{1.67}{mm} and \SI{0.0}{mm} root-mean-square errors in x- and y-direction, respectively. We further demonstrate that a surface area of 2.5~x~5\,mm is enough to read Braille letters with \SI{88}{\percent} average classification rate.

The great advantage of the acoustic sensing approach is that it effectively turns the whole actuator into a tactile sensor, without the need for dedicated tactile sensing hardware. Using only the embedded microphone and speaker, we can emulate not only sensors for contact location and force, object material, and actuator temperature (as shown previously), but also a tactile contact sensor on the whole actuator surface. This further manifests the great versatility and wide application range of the acoustic sensing approach.

\section{Related Work}
\label{sec:related-work}

In recent years, many promising new technologies for tactile sensing have been developed. A good overview of the different functioning principles can be found in the reviews by Zou et al.~\cite{zou_novel_2017}, Chi et al.~\cite{chi_recent_2018}, and Park et al.~\cite{park_recent_2018}. In this paper, we focus our attention on approaches that either measure contacts in form of tactile pixels (taxels), as well as those that employ acoustic sensing.

Camera-based sensors like GelSight~\cite{yuan_gelsight:_2017}, TacTip~\cite{ward-cherrier_tactip_2018}, and Insight~\cite{sun_soft_2022} obtain high-resolution tactile measurements by visually observing contact surfaces, which leaves the surface compliance of the sensorized object largely unaffected. However, the camera requires a line-of-sight connection to the point of contact, which makes it infeasible for continuum actuators with large deformations.

In our lab, we previously developed a tactile sensor using piezoresistive fabric and flex PCBs that is highly sensitive and easy to manufacture~\cite{pannen_low-cost_2022}. But it has a fixed number of taxels and is placed on the actuator's surface, which negatively affects its surface properties.
Other approaches use thin, flexible electronics to create 2D sensor arrays. Kaltenbrunner et al.\ created an ultra-lightweight sensor consisting of 12x12 taxels, each roughly \SI{6.6}{mm} wide~\cite{kaltenbrunner_ultra-lightweight_2013}.  
Nela et al.\ presented a design with 16x16 taxels, each circa 4mm wide~\cite{nela_large-area_2018}. 
And Wang et al.\ demonstrate a stretchable transistor array with a 10x10 taxel sensor with a width of \SI{2}{mm} each~\cite{wang_skin_2018}. 
Such solutions are very thin and highly flexible, which is ideal for soft surfaces like human hands and soft robots. However, the fabrication is difficult and the sensor patches require many electrical connections, resulting in a complex net of wires that would constrain soft actuators.


The second relevant research area is that of acoustic sensing. This promising sensing approach uses audio components to measure various different properties of sensorized objects, with high accuracy and minimal influence on the object's compliance behavior.
As Ono et al.\ and Harrison et al.\ have shown, sounds that travel through objects incur small variations in modulation  ~\cite{ono_touch_2013, harrison_tapsense_2011}. 
These variations in sound can be used to detect contact locations along air tubes~\cite{mikogai_contact_2020}, the bend angle of pneumatic actuators~\cite{randika_estimating_2021}, and grasping force during minimally invasive surgery~\cite{ly_grasper_2017}. Because many different properties affect the sound propagation, acoustic sensing can even be used for visual scene reconstruction~\cite{chen_boombox_2021} and temperature measurements on smartphones~\cite{cai_active_2021}. This demonstrates the large range of different properties that acoustic sensors can measure.


In our previous work, we have used these properties to sensorize pneumatic actuators to measure the contact location and contact forces, as well as the inflation level of the actuator, the material of objects the actuator touched, and the temperature of the actuator~\cite{wall_passive_2022}. Building upon these results, in this paper we demonstrate that the acoustic sensing approach can be extended to create a two-dimensional tactile array. We use the same sensor hardware and apply the acoustic sensing principle to measure small variations in complex contact patterns.

\section{Adapting the Acoustic Sensing Principle for 2D Tactile Arrays}
\label{sec:method}

\begin{figure}
\centering
\includegraphics[width=0.8\linewidth]{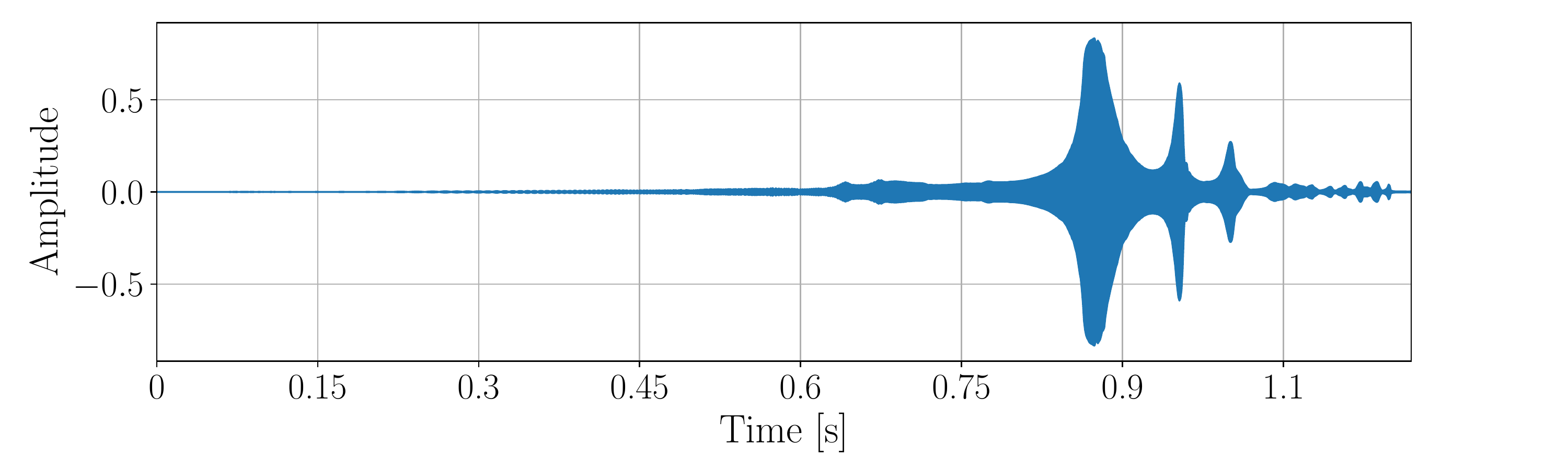}
\put(-210,47){a)}

\includegraphics[width=0.8\linewidth]{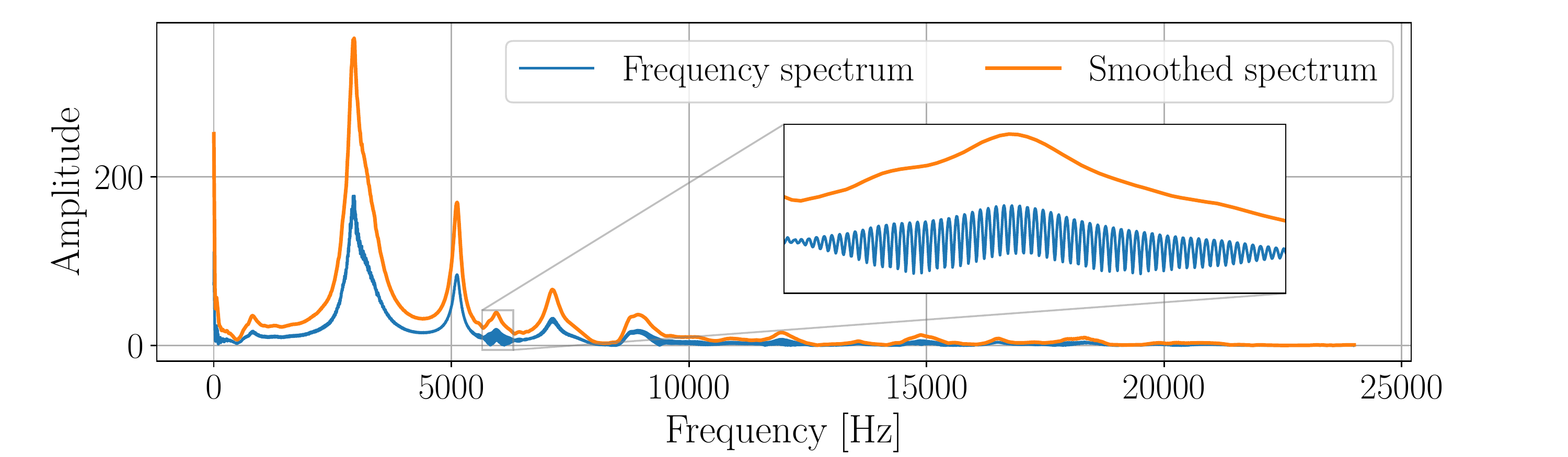}
\put(-210,47){b)}

\includegraphics[width=0.8\linewidth]{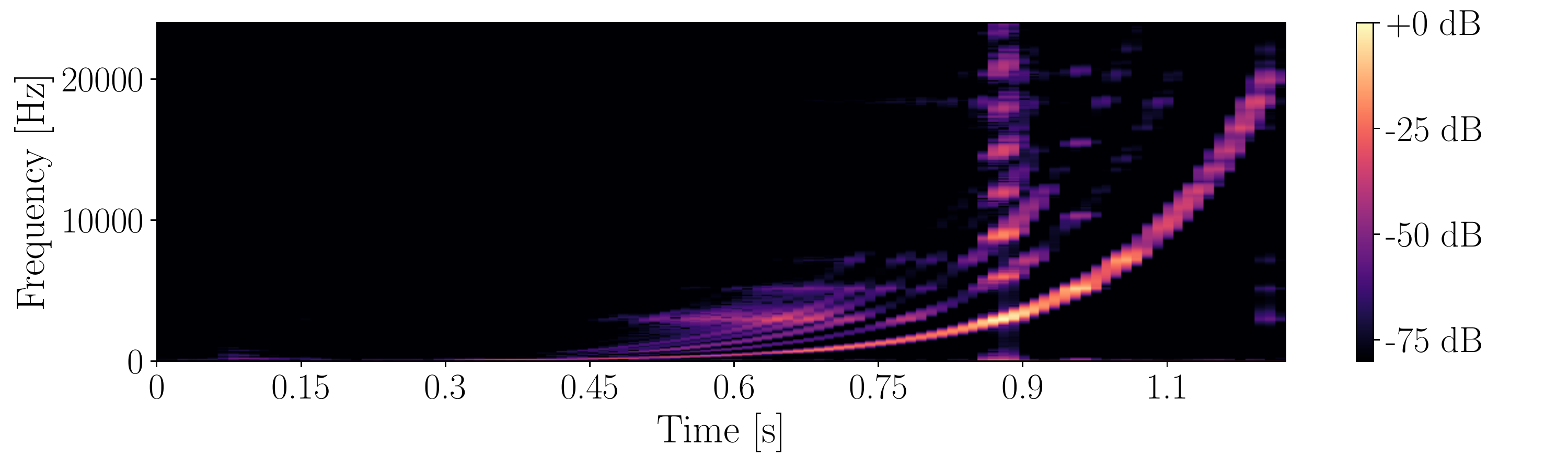}
\put(-210,47){c)}

\includegraphics[width=0.8\linewidth]{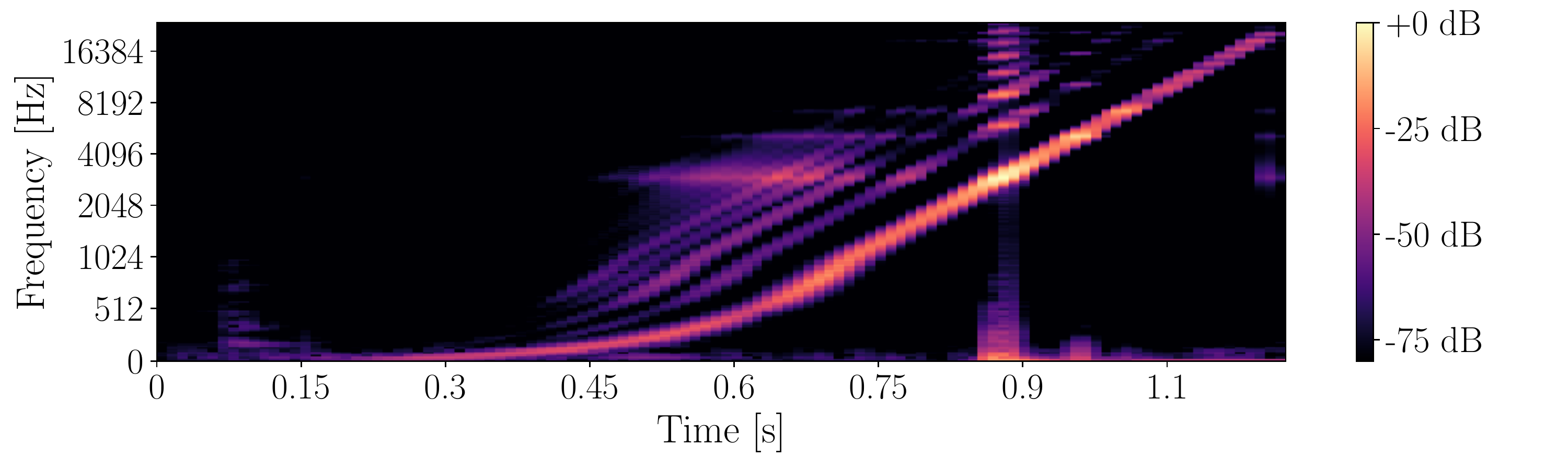}
\put(-210,47){d)}

\includegraphics[width=0.8\linewidth]{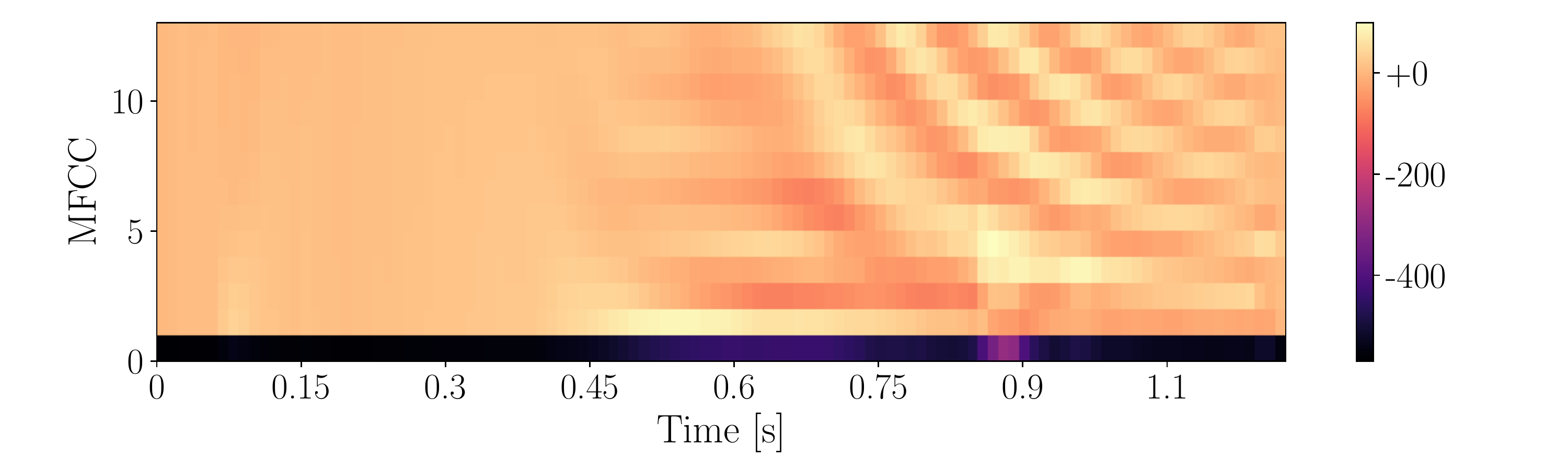}
\put(-210,47){e)}

\caption{Different sound representations of the same \SI{1}{\second} sweep audio recording: We compare the a) waveform, b) frequency spectra, c) spectrogram, d) Mel-scaled spectrogram, and e) Mel-frequency cepstral coefficients (MFCCs) to identify which works best to extract tactile information.}
\label{fig:soundrepresentations}
\end{figure}

In this paper, we present how to use acoustic sensing for tactile sensing on soft actuators. For this, we first summarize the general principle of acoustic sensing and then explain the key improvements that enable us to create a virtual, two-dimensional tactile sensing array with it.

\subsection{Acoustic Sensing Principle} 

Acoustic sensing is based on two key observations: 1.\ Sound that travels through structures gets modulated. 2.\ The exact modulation depends on the contact properties of the structure~\cite{cremer_structure-borne_2005}. Consequently, by detecting small variations in sound, we can learn to infer the contact property that caused that change in modulation. Practically, this can be achieved by generating a known sound with an embedded speaker, and recording the sound with an embedded microphone after the structure modulated it. Using data-driven supervised learning approaches, we then create sensor models that map sound recordings to corresponding contact states. But when attempting to use this approach to create a tactile sensing array, the challenge is that the differences in modulation can become quite small. Therefore, we need to carefully select a sound representation that maintains small differences, and a learning method capable of detecting them.

\subsection{Sound Representations} 
\label{sub:sound_representations}

The objective of selecting a sound representation is to capture all relevant details of the sound, while also making them easily accessible. However, due to the lack of reliable acoustic models for soft actuators~\cite{wall_passive_2022}, it is unclear which sound components carry the most information for acoustic sensing. Instead, we try out several different representations and select the one resulting in the best sensing performance.

Figure~\ref{fig:soundrepresentations} shows examples of the representations we evaluated. The ``waveform'' is directly what the microphone records. To visualize different resonance frequencies more clearly, the ``frequency spectrum'' is calculated using a fast Fourier transform. This is the representation we have previously used. However, we observed that data could be quite noisy and large. To address this, we calculate a ``smoothed spectrum'' by summing up the short-time Fourier Transform over time. This results in significantly less noise and fewer data points, without losing much detail.
Another commonly used sound representation for audio classification is the ``spectrogram''. It adds a time dimension to the spectrum and therefore captures how the frequencies change over time. A variant of this is the ``Mel-scaled spectrogram'' which uses the Mel scale to emphasize lower frequency ranges. Finally, the ``Mel-frequency cepstral coefficients''(MFCCs) represent the short-term power spectrum of the sound and are said to capture both linear and non-linear sound properties.

Comparing the tactile sensing performance of different sound representations, we found that the ``smooth spectrum'' performed best. We discuss our insights in Section~\ref{sec:discussion} and use the ``smooth spectrum'' for all reported results.

\subsection{Learning Methods} 
\label{sub:learning_methods}
We want to create sensor models that can detect and attribute specific sound modulations to the corresponding tactile event. We use a data-driven approach in that we use training samples to train or set up our model, which we then test on a separate, previously unseen set of test samples. Similar to the sound representations, it is not obvious which learning method is best suited to identify the relevant modulation differences. So again we evaluate a range of different methods: 
k-nearest neighbors (KNN), support-vector machines (SVM), basic fully-connected neural networks (NN), and convolutional neural networks (CNN). 
KNNs are simple and fast but weight equally the whole feature vector. SVMs can learn class boundaries, but the choice of kernel function is challenging. NNs are very powerful, but require good tuning and many training samples. CNNs might work well with the ``image''-like properties of the spectrogram representations. But they tend to ignore the temporal order of the spectrogram image and have many parameters to tune. 
We implemented the KNN and SVM models using the scikit-learn framework\footnote{\url{https://scikit-learn.org}} and the NN and CNN models using PyTorch\footnote{\url{https://pytorch.org}}. 

\section{Experimental Validation}
\label{sec:experiments}
So far we have explained the acoustic sensing principle and how to adapt it to two-dimensional tactile sensing. We now demonstrate our approach in practical experiments. For this, we use a PneuFlex actuator~\cite{deimel_novel_2016} and a programmable Braille display (Fig.~\ref{fig:braillesetup}). 
The soft materials of this pneumatic continuum actuator make it inherently flexible, which makes traditional, ``hard'' sensors infeasible. Additionally, the actuator's air chamber provides sufficient space to embed the audio components. 
The Braille display consists of 32~x~4 individually controllable pins which allows us to easily create different contact patterns on the surface of the actuator with low delay and high repeatability.

\subsection{Experimental Setup Using a Braille Display} 
The fabrication steps of the ``Acoustic Sensing'' PneuFlex actuator remain identical to our previous publications~\cite{zoller_active_2020}: We attach a MEMS condenser microphone (Adafruit SPW2430) at the base of the actuator's air chamber, and a balanced armature speaker (Knowles RAB-32063-000) at the tip. Both components have a similar linear response range of \SI{100}{Hz}--\SI{10}{KHz}. The speaker emits a \SI{1}{\second} frequency sweep. The placement on opposite ends of the air chamber maximizes the travel distance of the sound, increasing the contact-dependent modulation effects.
A USB audio interface (MAYA44 USB+) drives both audio components at a sample rate of \SI{48}{kHz} with \SI{32}{bit} precision. 
The finished actuator has a palmar surface area of approximately 90~x~10\,mm.

The Braille display consists of two stacked modules with eight Braille cells each (Metec AG, Flat PCB 8) for a total of 32x4 pins on an area of 103x10\,mm. Between two cells the space is approximately \SI{6.42}{mm}. Within each cell, the pins have a spacing of \SI{2.45}{mm}. Each pin has an extended height of \SI{0.7}{mm} and a lifting force of \SI{0.17}{N}. We use a RapsberryPi to control each pin individually. 

We mount the sensorized PneuFlex actuator on top of the Braille display. The \SI{90}{mm} long actuator overlaps with 15 of the 16 Braille cells. Due to the manual fabrication, the actuator's palmar surface is not completely even. To ensure pin contact everywhere, we apply a small downward force during the mounting so that the whole surface makes contact with the Braille display.
All recorded sound samples are converted into smoothed spectra, split into training and test sets, and finally used to train and evaluate the sensor models.

\subsection{Resolution Analysis in X- And Y-Direction Shows Root-Mean-Square Errors of \SI{1.67}{mm} And \SI{0.0}{mm}} 
We start by analyzing the spatial resolution of the acoustic tactile sensor array in x and y-direction. Ideally, it would identify the exact location of contact in both dimensions. 

Additionally, this demonstrates another key feature of using acoustics for tactile sensing: Because the sensor hardware is not made up of \emph{physical} taxels, we can effectively change the shape of the \emph{virtual} taxels by training on a different dataset. We demonstrate this here with two separate sensor models for detecting edges in x and y-direction. 

We use the Braille display to record two datasets: One with 29 lines of pins along the x-dimension, and another one with 4 lines of pins along the y-dimension. The exact distance between two lines is determined by the pin-spacing of the Braille display of \SI{2.45}{mm} within cells and \SI{3.97}{mm} between cells. For the 29 x-lines, we record 200 samples per class and for the 4 y-lines, we record 125 samples per class. Data are split 3:2 into training and test samples. For both datasets, we perform a separate grid-search for a k-nearest-neighbor-regression model.

\begin{figure}
	\centering
	\subfigure[Example pattern of x-line]{
		\includegraphics[width=0.7\linewidth]{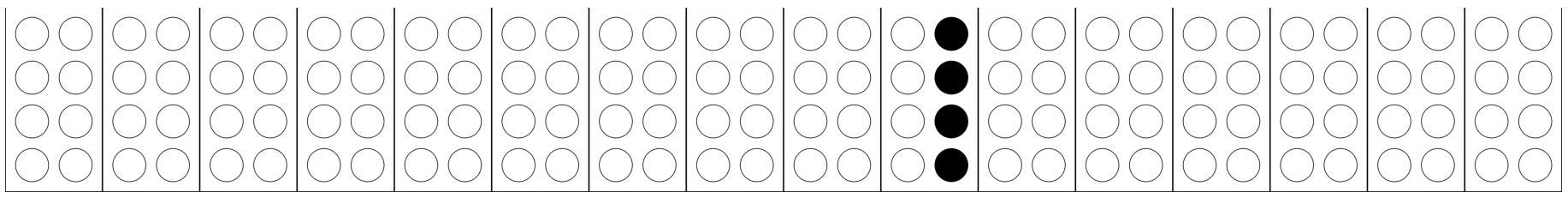}
		}
	\subfigure[Example pattern of y-line]{
		\includegraphics[width=0.7\linewidth]{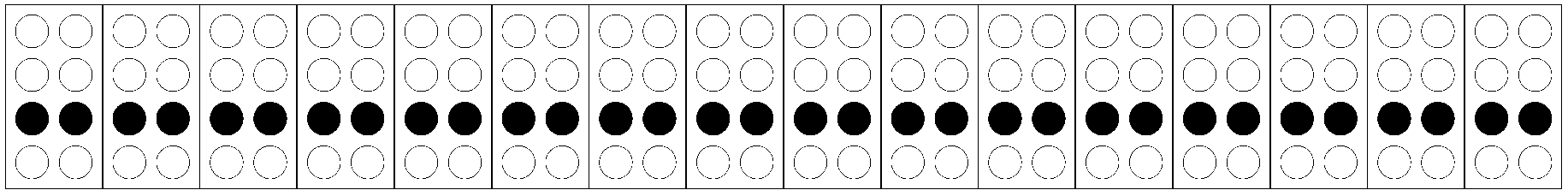}
		}
	\subfigure[Regression of x-position]{
		\includegraphics[width=0.9\linewidth,trim={0.2cm, 0.7cm, 0.7cm, 1.1cm},clip]{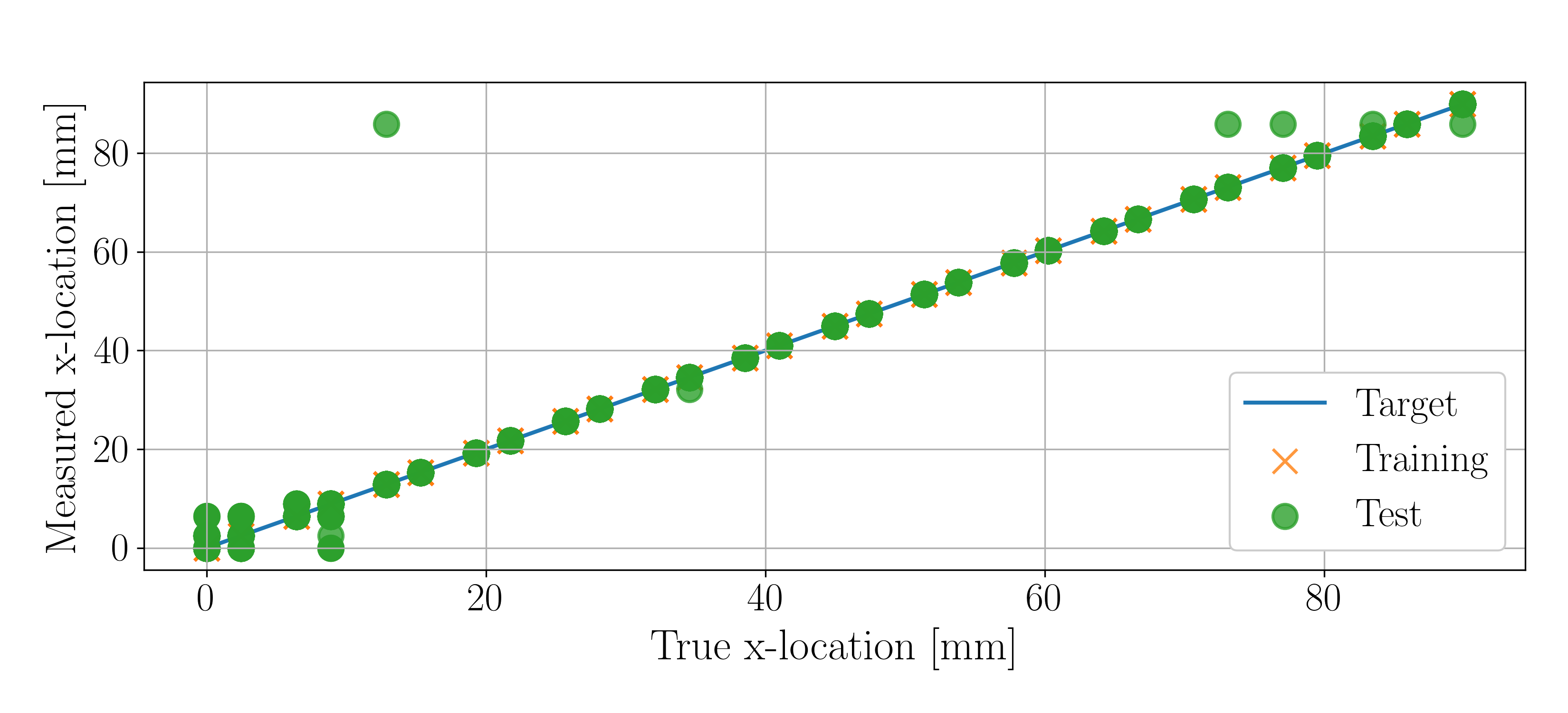}
		}
	\subfigure[Regression of y-position]{
		\includegraphics[width=0.9\linewidth,trim={0.2cm, 0.7cm, 0.7cm, 1.1cm},clip]{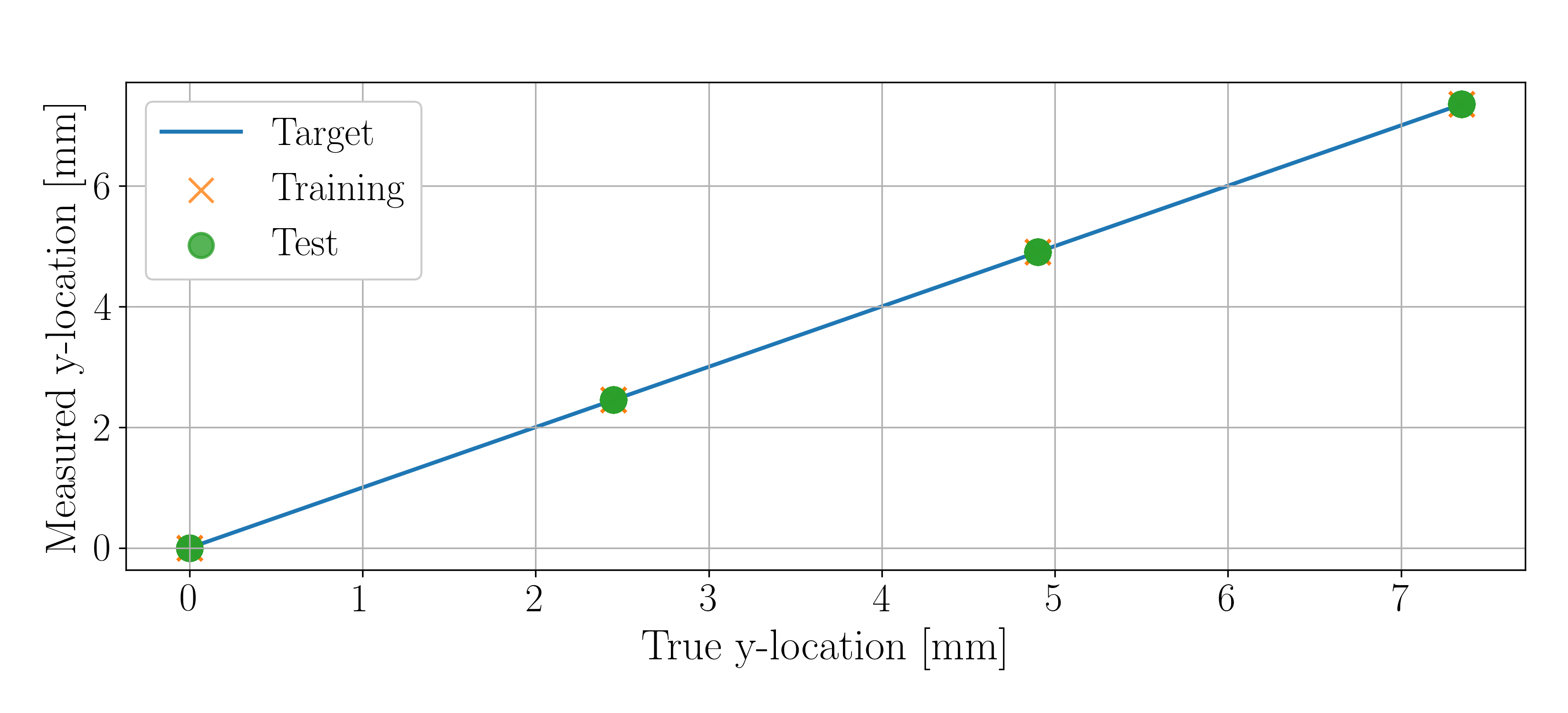}
		}
	\caption{Spatial resolution in the x- and y-direction of the tactile array: Using a Braille display to create line patterns with a spacing of \SI{2.45}{mm}, the acoustic sensor achieves an RMSE of \SI{1.67}{mm} in the x-direction and a perfect RMSE of \SI{0.0}{mm} in the y-direction.
	}
	\label{fig:exp_x_y_regression}
\end{figure}

The regression plots in Figure~\ref{fig:exp_x_y_regression} show the high accuracy of the acoustic sensor. Along the longer x-dimension, the root-mean-square error (RMSE) across the 2320 test samples is \SI{1.67}{mm}. And in the shorter y-dimension, all 200 test samples were predicted exactly for an RMSE of \SI{0}{mm}. 
In addition to demonstrating high sensing resolution, this also highlights how easily the ``shape'' of our virtual sensing taxels can be reconfigured, simply by using a different dataset when training the sensor model.

\subsection{Sensitivity Analysis Shows High Accuracy Even For Low-Force Contacts}
\label{sec:single_double_pin} 
Next, we evaluate the sensitivity of the 2D tactile array to light contacts. This is potentially difficult for an acoustic sensor, as light contacts likely result in only small changes to the actuator's sound modulation properties. We test this with a single point of contact and calculate the distance between true and measured contact location on the 2D tactile surface. Additionally, this will tell us if there are regions of the surface that offer different sensing accuracy.

\begin{figure}
	\centering
	\subfigure[Example pattern with a single pin]{
		\includegraphics[width=0.7\linewidth]{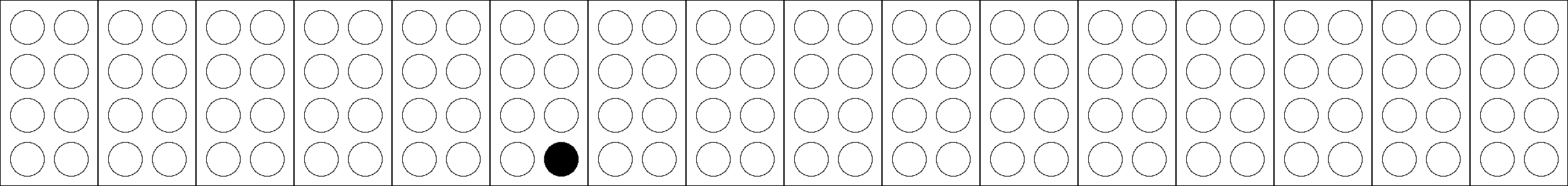}
		}
	\subfigure[Example pattern with a 2x2 patch]{
		\includegraphics[width=0.7\linewidth]{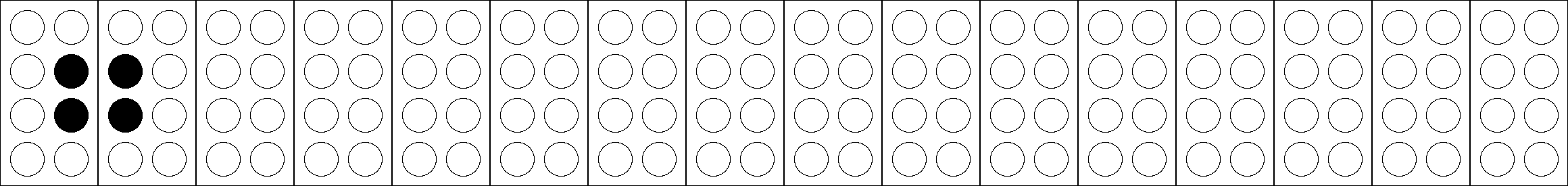}
		}
	\subfigure[Accuracy in x- and y-Direction]{
		\includegraphics[width=0.9\linewidth, trim={0cm, 0.2cm, 0.7cm, 1cm}, clip]{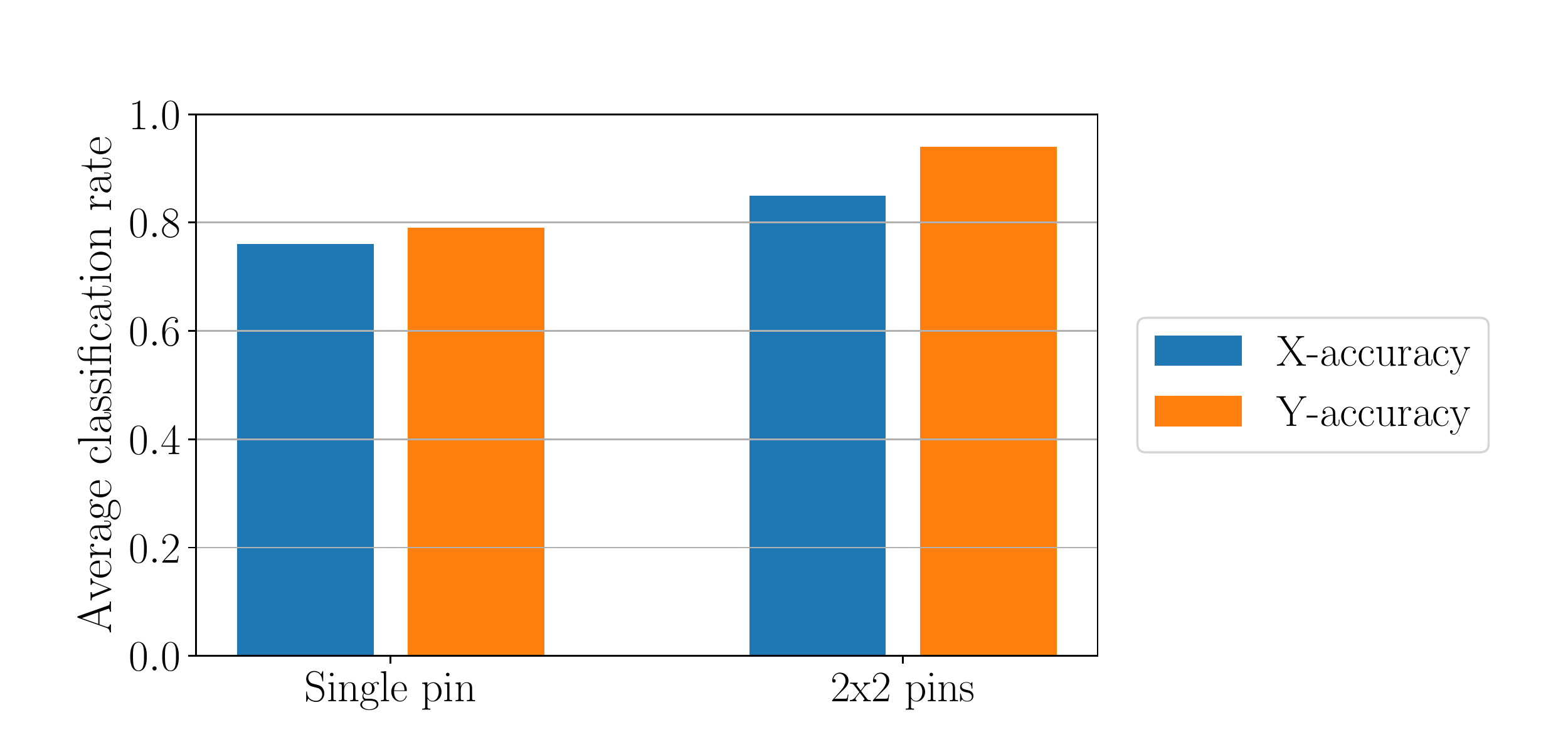}
		}
	\subfigure[Single pin error distribution]{
		\includegraphics[width=1\linewidth]{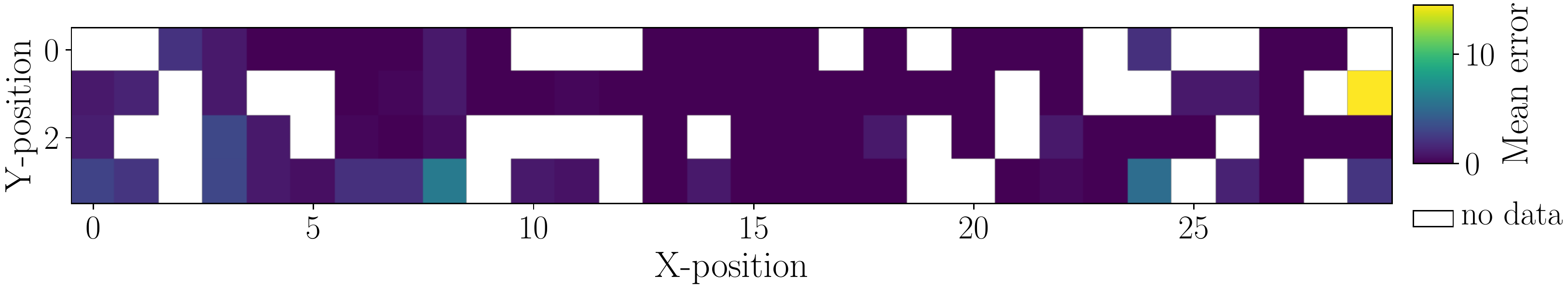}
		\label{subfig:singlepixelerror}
		}		
	\subfigure[2x2 pin error distribution]{
		\includegraphics[width=1\linewidth, trim={0.8cm 0.5cm 0.3cm 0.6cm}, clip]{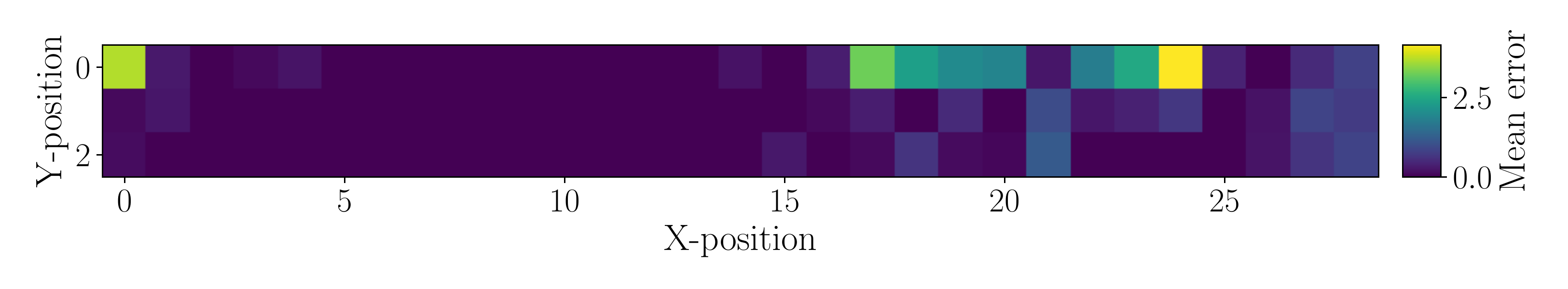}
		\label{subfig:doublepixelerror}
		}
	\caption{Sensitivity analysis using single pins and 2x2 pin patterns: A single pin of the Braille display has only a small lifting force of \SI{0.17}{N}. Nonetheless, the acoustic tactile array achieves an average classification rate of \SI{76}{\percent} and \SI{79}{\percent} in x and y-direction, respectively. Using the larger 2x2 pin pattern, the prediction rate is increased to \SI{85}{\percent} and \SI{94}{\percent}. The bottom two plots show the mean error distance for each taxel across the tactile array (blank taxels had no samples in the test data set). The average error distance for single and 2x2 pin sensing are 0.80 and 0.39 taxels, respectively.
	}
	\label{fig:exp_single_double_pins}
\end{figure}

To generate the dataset, we again use the Braille display to make contact with the actuator at known locations. A single pin of the Braille display has a lifting force of only \SI{0.17}{N}. For the first dataset, we record 500 randomly sampled contact points, each with a single pin extended. To test if higher contact forces improve sensing accuracy, we additionally record a second dataset of 2000 random samples using a 2x2 pin pattern to increase the pushing force and size of the contact patch (see Figures~\ref{fig:exp_single_double_pins}a and b). We split both datasets 2:1 into training and test sets and perform grid searches for two support-vector-classifier models. 

Figure~\ref{fig:exp_single_double_pins} shows the average classification rates for the ``single pin'' and ``2x2 pins'' cases, indicating the percentage of \emph{exactly} correct measurements split into the x- and y-direction. The measurement of the x-position (blue) can be seen to be slightly worse than the y-position (orange) in both cases. This matches the findings in the previous section. Nevertheless, the data shows that in \SI{76}{\percent} and \SI{79}{\percent} of samples, the x- and y-position, respectively, are correctly identified given only a single pin of the Braille display making contact. As expected, the x- and y-accuracy increases further to \SI{85}{\percent} and \SI{94}{\percent}, respectively, when the contact patch consists of a 2x2 pin pattern. 

Additionally, Figures~\ref{subfig:singlepixelerror} and~\ref{subfig:doublepixelerror} visualize the average error distance for each virtual taxel for the single and 2x2 pin dataset. Low values (dark blue) indicate good measurements as the predicted contact point is close to the true contact point. This is the case for most of the sensor area, which demonstrates high accuracy and mostly uniform distribution. Large values (bright green), like in the top row, reveal that measurements in this area are less precise. We believe that these measurements can be improved by a) ensuring a good, even contact between actuator and braille display and b) recording additional samples for less accurate regions.

\subsection{Reading Braille Letters with \SI{88}{\%} Accuracy} 
Finally, we demonstrate the sensor's ability to differentiate between complex contact patterns: the Braille alphabet. This shows that even very small changes in the contact's shape result in a detectable difference in sound modulation.

\begin{figure}
	\centering
	\subfigure[The Braille alphabet]{
		\includegraphics[width=0.7\linewidth]{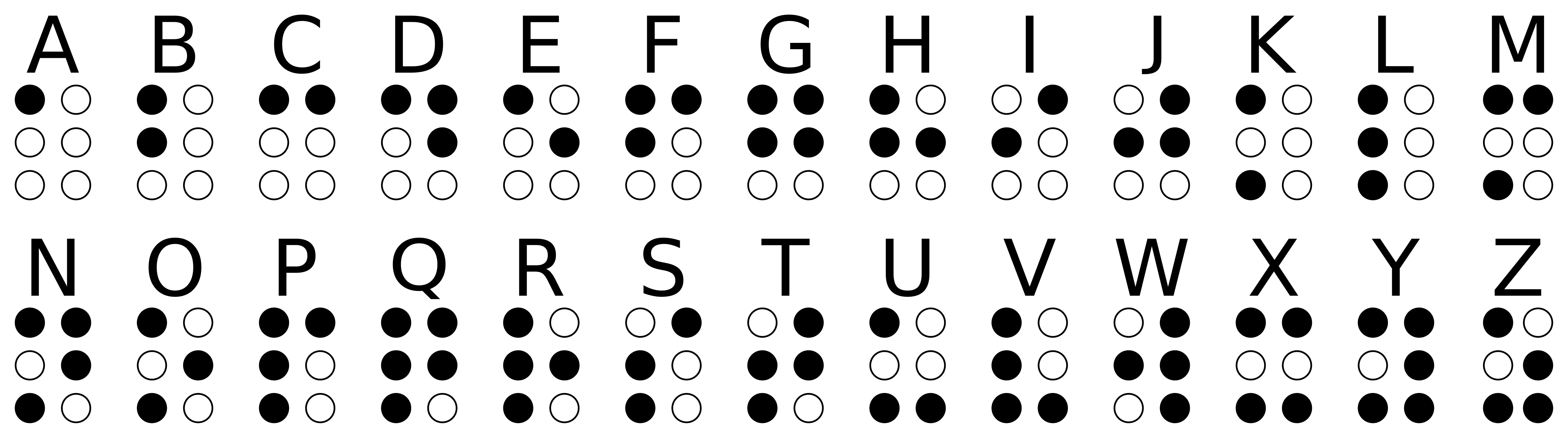}
		\label{subfig:braille_alphabet}
		}
	\subfigure[Example of Braille display for the letter 'z' ]{
		\includegraphics[width=0.7\linewidth]{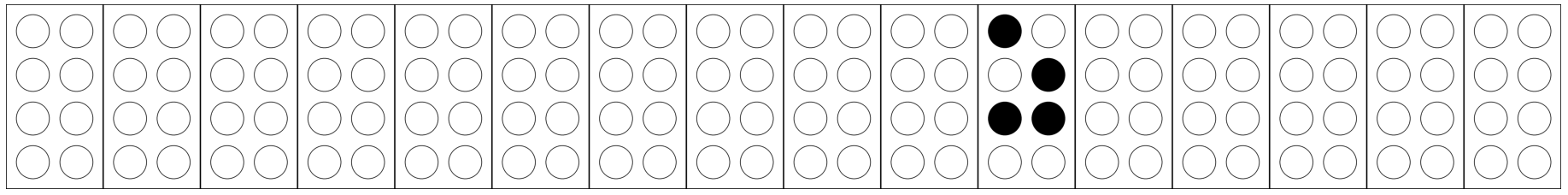}
		\label{subfig:braille_chars_examples}
		}
	\subfigure[Confusion matrix for Braille letter classification]{
		\includegraphics[width=\linewidth, trim={0, 1.5cm, 0, 1.7cm}, clip]{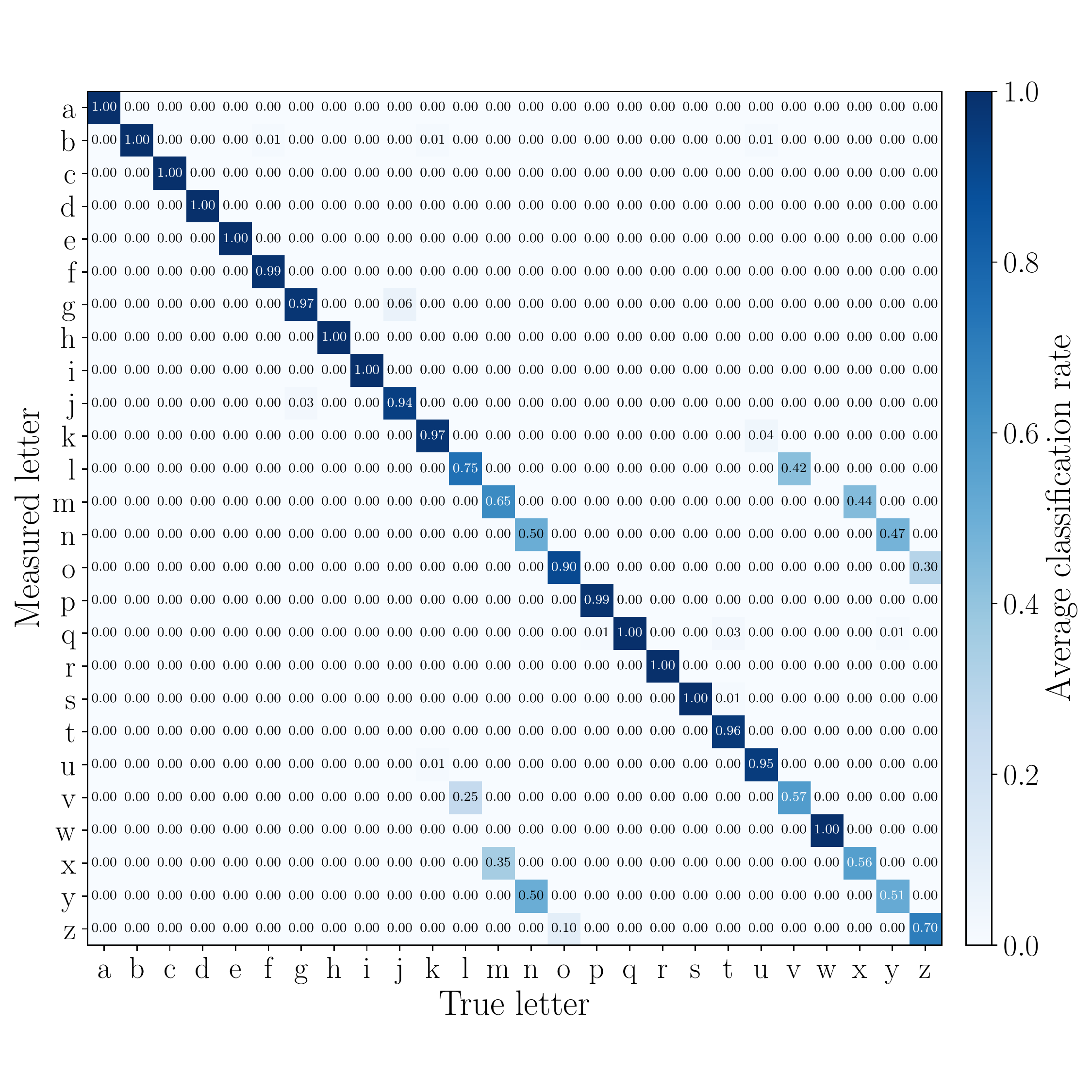}
		\label{subfig:cm_braille_chars}
		}
	\caption{Reading Braille letters: We use a single cell of the Braille display with a size of only 2.5~x~5\,mm to show all 26 letters of the alphabet. The acoustic tactile array successfully recognized the complex patterns with an average classification rate of \SI{88}{\percent}. The large values on the diagonal of the confusion matrix demonstrate the sensor's high accuracy with only a few letters being mixed up.
	}
	\label{fig:exp_braille_characters}
\end{figure}

For this experiment, only a single Braille cell is used to display each letter of the alphabet in an area of approximately 2.5~x~5\,mm. All other pins remain retracted (Fig.~\ref{subfig:braille_chars_examples}). We record a dataset with 200 samples per letter and split the resulting 5200 total samples 3:2 into training and test set with an equal class distribution. We use a support vector classifier and perform a grid search over its hyper-parameters.

\begin{wrapfigure}[18]{r}{0.5\linewidth}
	\centering
	\vspace{-2ex}
	\includegraphics[width=0.9\linewidth, trim={0.5cm, 0.8cm, 0, 1.2cm}, clip]{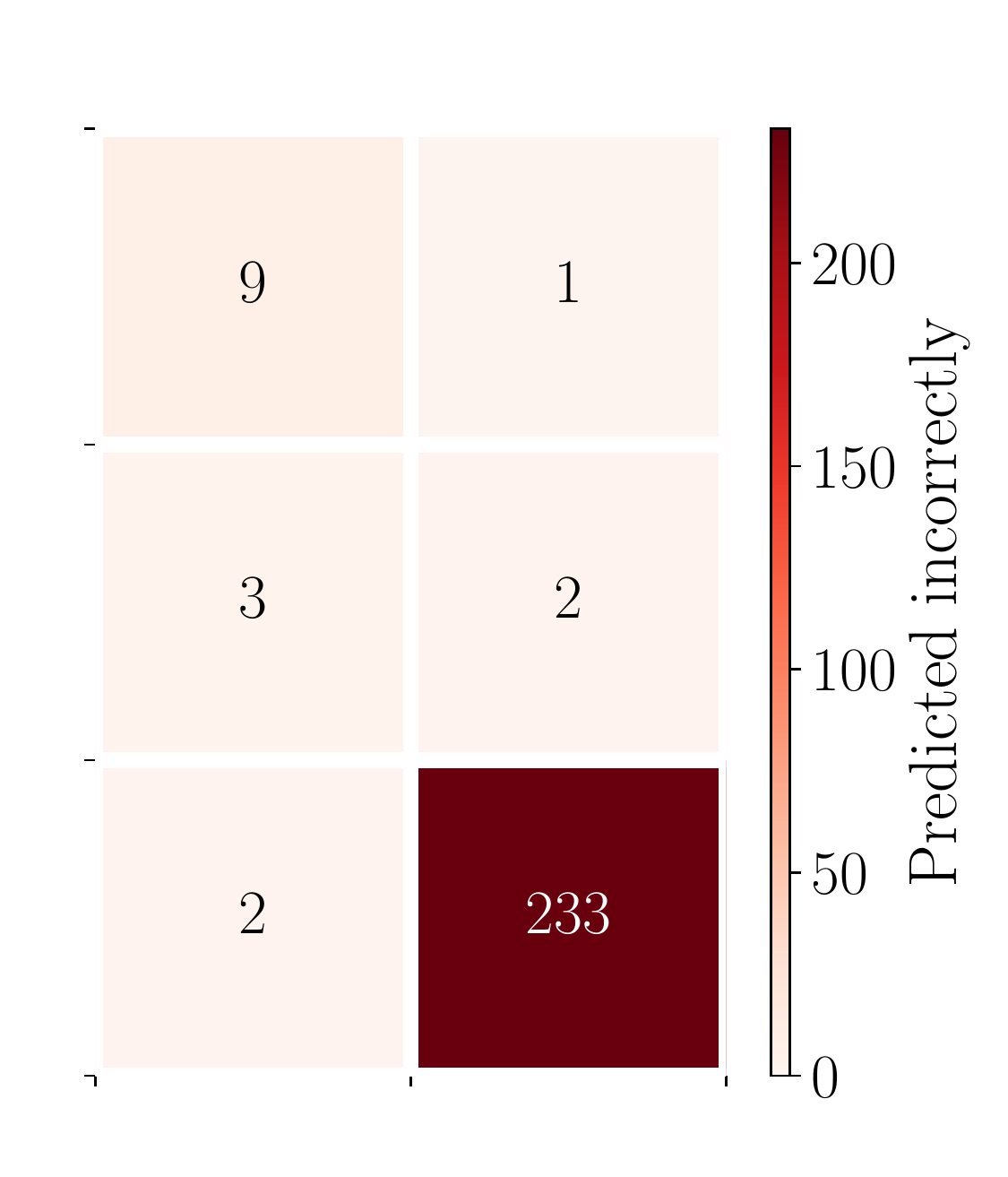}
	\caption{When comparing the pins of misclassified Braille letters, it becomes obvious that the pin in the bottom right is causing the most errors. This is also the pin that is used by the fewest letters (u-z). Additional training samples may help to correct this shortcoming.}
	\label{subfig:wrong_pins}
\end{wrapfigure}

Figure~\ref{subfig:cm_braille_chars} shows the confusion matrix for the 2080 test samples. The high values on the diagonal demonstrate the very reliable measurements of Braille letters using only the 2.5~x~5\,mm area of the acoustic tactile array. The average classification rate of letters is \SI{88}{\percent}. 
In Figure~\ref{subfig:wrong_pins}, we show in which pins the misread Braille letters differed. It can be seen, that the pin in the bottom right corner is involved in most misclassifications. This also explains the four ``pairs'' of often-confused letters: l/v, m/x, n/y, and o/z. For each pair, the only difference is the bottom right pin. 

Taking the idea of reading Braille with the acoustic tactile sensor even a bit further, we simulated the reading of a text made up of the \num{3000} most common English words\footnote{\url{https://www.ef.com/wwen/english-resources/english-vocabulary/top-3000-words/}}. We randomly sampled \num{100000} words and applied the probabilities from the confusion matrix in Fig.~\ref{subfig:cm_braille_chars} to simulate the misclassification of individual letters. 
We then used a very simple heuristic based on the Hamming distance of words, to find the most likely word candidate. Using this strategy, the tactile sensor could identify the correct word \SI{95}{\percent} of the time. In less than \SI{1}{\percent} a measurement error results in a different existing word, in \SI{5}{\percent} the heuristic fails and returns a wrong word, but in \SI{39}{\percent} the simple heuristic successfully corrects a misread word back to the original.


In any case, the classification rate of \SI{88}{\percent} for reading Braille letters in a sensor area of 2.5~x~5\,mm demonstrates that the acoustic tactile sensor array measures even complex contact patterns of Braille letters with high accuracy.

\section{Discussion}
\label{sec:discussion}
\begin{figure}
\centering
\includegraphics[width=1\linewidth, trim={0.7cm, 1.2cm, 0.4cm 2.4cm}, clip]{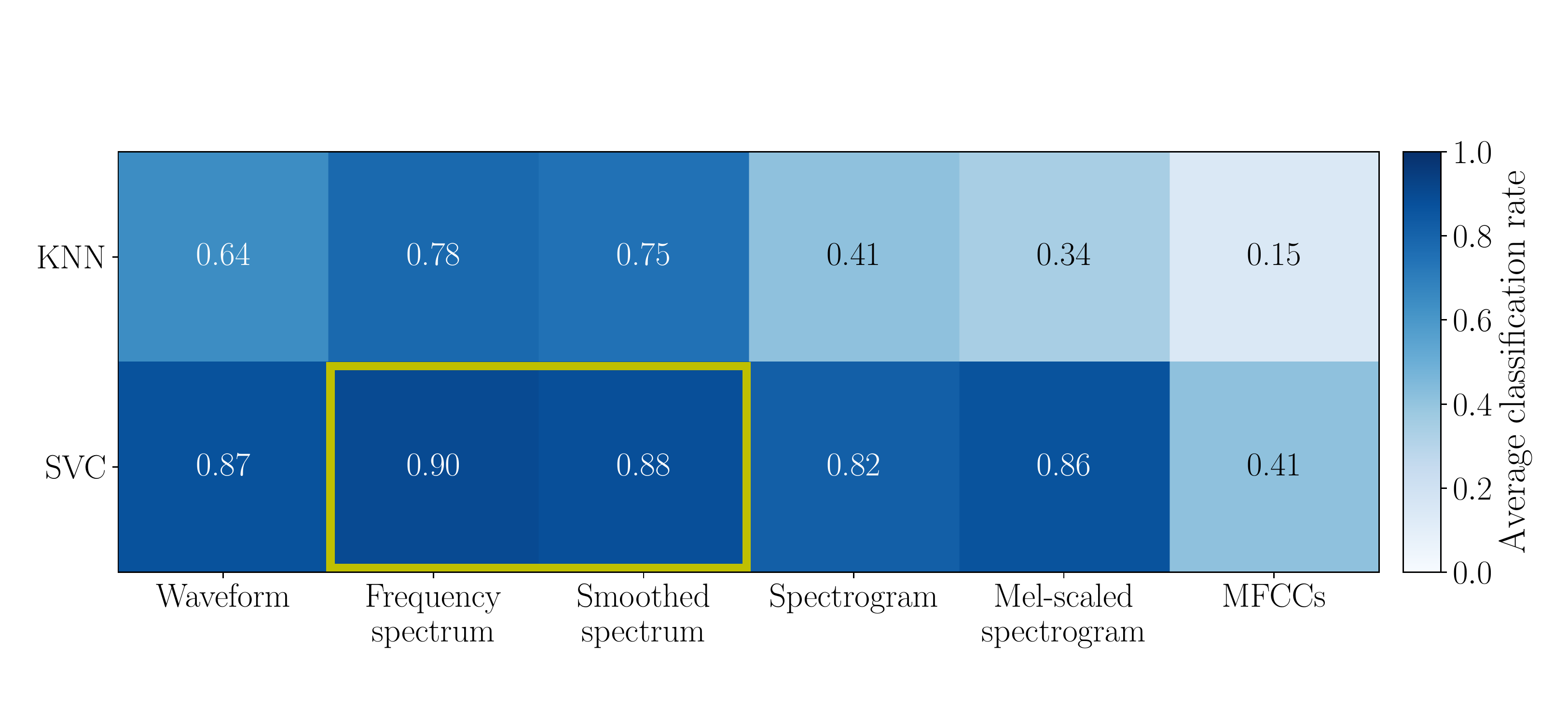}
\caption{When comparing the classification accuracy of Braille characters, the support vector classifier shows good results for most sound representations, with the \emph{regular} and \emph{smoothed} frequency spectra achieving the highest values of \SI{90}{\percent} and \SI{88}{\percent}, respectively. }
\label{fig:comparereprmodels}
\end{figure}

The evaluation of different sound representations (Fig.~\ref{fig:comparereprmodels}) yielded a surprising result: The two comparatively simple \emph{frequency spectra} outperformed the other (more complex) representations, including those with additional information via the temporal axis, e.g.~the spectrogram. We believe that this is due to the complexity of the passive shape adaptation of the soft actuators which in turn results in distinct changes in the sound modulation. The actuator's compliance effectively ``embeds'' the relevant tactile information into the sound, making it unnecessary to use more complex representations. We chose to use the \emph{smoothed} spectrum, as it performs similar to the \emph{regular} spectrum, but is more robust and significantly speeds up learning. 

When comparing the different learning methods, we observed that the simple KNN classifier worked well for simpler problems like edge localization. But SVM achieved the best results for complex problems like Braille character recognition. We were unable to achieve comparable results with the network-based methods, which may be due to the limited number of training samples or our limited experience with tuning network parameters. Neural nets have recently been shown to perform well with contact microphones~\cite{chen_boombox_2021}, so they may still be useful for sensing problems where the simple models fail to achieve good results.

Finally, it is worth noting that the experiments in this paper aim to show the \emph{possibility} to use acoustic sensing to create accurate virtual tactile arrays. While the reading of a Braille display may have limited practical applications, it demonstrates that the acoustic sensor can extract detailed tactile information from the surface of a soft actuator. Next, we plan to investigate the sensing behavior with novel objects and situations for practical use of the acoustic tactile sensor with a dexterous soft hand, like the RBO Hand~3~\cite{puhlmann_rbo_2022}.

\section{Conclusion}
We presented a virtual two-dimensional tactile array based on acoustic sensing. Audio components embedded into a soft actuator record contact-depend changes in internal sounds and a trained sensor model maps these sound changes to contact patterns. We discussed different sound representations and learning methods to improve spatial resolution and sensitivity to even small contact events. 

We demonstrated the acoustic tactile array on the soft PneuFlex actuator using a Braille display. We determined the spatial resolution of the sensor in x- and y-direction to be close to the 2.45\,mm pin-spacing of the Braille display, with root-mean-square regression errors of \SI{1.67}{mm} and \SI{0.0}{mm}, respectively. Even for light contacts of a single Braille display pin with a force of \SI{0.17}{N} the x- and y-position were correctly classified with \SI{76}{\percent} and \SI{79}{\percent} accuracy.
Finally, we showed that the acoustic tactile array could read all 26 letters of the Braille alphabet from a single Braille cell with a classification rate of \SI{88}{\percent}.


We believe these results demonstrate the impressive detail achievable when using sound. The simple hardware and large range of applications make acoustic sensing a highly versatile approach for tactile sensing in soft robotics.


\balance


\bibliographystyle{IEEEtran}
\bibliography{acoustictaxels2022}

\end{document}